\documentclass[a4paper, 10pt, twocolumn, twoside]{article}

\usepackage{Improvisational}

\usepackage{lscape}

\usepackage{hologo}

\usepackage{array}

\usepackage{tabularx}

\usepackage{booktabs}

\usepackage{booktabs}

\usepackage{graphicx} 

\usepackage{multirow}

\usepackage[normalem]{ulem}

\usepackage{url}

\usepackage{hyperref}

\usepackage{dblfloatfix}

\newcolumntype{Y}{>{\centering\arraybackslash}X}

\begin{document}

\linespread{0.5}

\setlength{\parskip}{0.5em}

\title{ Task-Aware Positioning for Improvisational Tasks in Mobile Construction Robots via an AI Agent with Multi-LMM Modules}

\author{Seongju Jang$^{a}$, Francis Baek$^{b}$, SangHyun Lee$^{c}$}

\affiliation{
\parbox{1\linewidth}{
\raggedright
$^a$Ph.D. Student, Department of Civil and Environmental Engineering, University of Michigan, 2350 Hayward St, Ann Arbor, MI 48109, E-mail: jseongju@umcih.edu\\
$^b$Assistant Professor, School of Civil and Environmental Engineering, Georgia Institute of Technology, North Avenue, Atlanta GA 30332, E-mail: francis.baek@ce.gatech.edu\\
$^c$Robert B. Harris Collegiate Professor, Department of Civil and Environmental Engineering, University of Michigan, 2350 Hayward St, Ann Arbor, MI 48109, E-mail: shdpm@umich.edu (corresponding author)\\
}
}

\maketitle 
\thispagestyle{fancy} 
\pagestyle{fancy}

\begin{abstract}
Due to the ever-changing nature of construction, many tasks on sites occur in an improvisational manner. Existing mobile construction robot studies remain limited in addressing improvisational tasks, where task-required locations, timing of task occurrence, and contextual information required for task execution are not known in advance. We propose an agent that understands improvisational tasks given in natural language, identifies the task-required location, and positions itself. The agent’s functionality was decomposed into three Large Multimodal Model (LMM) modules operating in parallel, enabling the application of LMMs for task interpretation and breakdown, construction drawing-based navigation, and visual reasoning to identify non-predefined task-required locations. The agent was implemented with a quadruped robot and achieved a 92.2\% success rate for identifying and positioning at task-required locations across three tests designed to assess improvisational task handling. This study enables mobile construction robots to perform non-predefined tasks autonomously.
\end{abstract}


\section{Introduction}
\label{sec:Introduction}

Stagnant productivity growth and labor shortages are critical challenges in the modern construction industry \cite{ref_1}. From 2000 to 2022, global construction productivity grew by only 0.4\% annually, compared to 3\% in manufacturing, underscoring the construction industry’s persistently low productivity growth \cite{ref_2}. In addition, according to the 2022 Workforce Survey \cite{ref_3}, 91\% of construction firms with open positions struggled to fill those roles, highlighting the chronic labor shortage in the construction industry. 

The adoption of autonomous robots has the potential to address the labor shortages and low productivity growth in the construction industry \cite{ref_4, ref_5}. Autonomous robots can continuously perform repetitive and physically demanding tasks without rest, leading to improved productivity and reduced labor demands \cite{ref_6}. Although autonomous robots can exist in various operational forms, robots equipped with mobility show particularly strong potential for performing tasks in construction environments \cite{ref_4}. Because many construction tasks are not performed at a single fixed location, mobility allows robots to be easily deployed across multiple task-required locations on construction sites \cite{ref_4}. Accordingly, many construction robots with mobility based on wheeled, tracked, and legged platforms have been developed and deployed on construction sites \cite{ref_4, ref_7, ref_8, ref_9, ref_10}.

Mobile construction robots are expected to improve productivity and reduce labor demand across the majority of construction activities, including both direct and auxiliary works. Direct work refers to construction work that directly adds value to the structure itself \cite{ref_11}. Several mobile robots have already been tested in direct work tasks, such as overhead drilling and drywall finishing, reducing human labor and demonstrating improvements in productivity \cite{ref_9, ref_10}. Auxiliary work refers to activities that support direct work, including tasks of site preparation, material handling, tool setup, and monitoring. According to on-site activity analysis, auxiliary work accounts for a substantial 47.8\% of productive work time on construction sites, a proportion nearly similar to the direct work (52.2\%) \cite{ref_12}. Mobile robots have also been investigated for performing these auxiliary work tasks, such as layout drawing, material handling, and indoor safety inspection, reducing both required labor and execution time \cite{ref_10, ref_13, ref_14}.

While showing potential in many tasks, for mobile construction robots to meaningfully drive productivity gains and labor reductions on ever-changing construction sites, the ability to handle tasks that arise improvisationally is essential. Tasks in construction work may be carried out according to predefined plans, but are also often performed improvisationally, meaning that task instructions, execution locations, and timing can vary depending on site conditions \cite{ref_15}. Direct work frequently requires improvisational adjustments from the plan in response to site conditions \cite{ref_15}, whereas auxiliary work is even more commonly performed on an as-needed basis rather than following a predefined, detailed task plan \cite{ref_16}. Mobile robots hold significant potential to enhance productivity and reduce labor demand across both direct and auxiliary work. Yet, given the prevalence of improvisation in construction, the practical contribution of mobile robots can be largely constrained if they cannot reliably handle these improvisational tasks.

However, the ability of mobile construction robots to autonomously position themselves at task-required locations for improvisational tasks remains largely unexplored. For improvisational tasks, task instruction, execution location, and task assignment or modification timing are not predefined \cite{ref_15, ref_17}. Therefore, robots must be capable of accepting and interpreting task instructions when they are given, identifying the required task target, and positioning themselves to it. In particular, in complex construction sites, it is often necessary to identify the intended task target by considering not only the object itself but also the contextual conditions (e.g., spatial relations) embedded in the command and the attribute conditions (e.g., shape, size, color) of the task target. Although recent studies on Large Multimodal Model (LMM)-based navigation have shown potential for robots to handle improvisation by understanding non-predefined commands \cite{ref_18, ref_19, ref_20, ref_21, ref_22, ref_23}, these approaches remain insufficient for improvisational tasks on construction sites. Many LMM-based approaches still assume that task-required locations are predefined \cite{ref_21, ref_22, ref_23}. While some recent approaches identify previously unknown targets and navigate to them, the selection of an appropriate task-required location under non-predefined attribute-based or contextual conditions has not been addressed \cite{ref_18, ref_20}. Furthermore, the ability to flexibly assign or modify tasks during execution remains limited \cite{ref_18, ref_19, ref_20}.

In this research, we established an agent for construction robots that can flexibly respond to improvisational tasks on construction sites, enabling autonomous identification and positioning at task-required locations for each task. Specifically, we proposed an agent with multiple LMM modules that addresses three key challenges in autonomous positioning for improvisational tasks: (1) identifying task targets whose exact locations are unknown; (2) handling task targets with non-predefined attributes or contextual conditions; and (3) flexibly accommodating new tasks that are assigned or modified anytime during execution. We implemented our agent with a quadruped robot, Unitree Go2 \cite{ref_24}. The agent was realized using a Docker-based backend architecture that allows three modules to operate in parallel and to provide real-time feedback \cite{ref_25}. We designed three tests in a controlled indoor environment to demonstrate the three challenges of improvisational tasks on construction sites and to evaluate how successfully the robot positioned itself at task-required locations in response to them. 

The proposed agent is capable of positioning the construction robot at task-required locations for improvisational tasks on construction sites, even when task-required locations, task instructions, including attribute and contextual conditions, and the timing of task assignment or modification are not known in advance. This study enables mobile construction robots to autonomously perform diverse improvisational tasks that constitute a substantial portion of construction work.

\section{Literature review on autonomous robot positioning}
\label{sec:LiteratureReview}

In construction sites, where tasks are not generally confined to fixed locations, mobility has been recognized as an important capability for construction robots \cite{ref_4}. For construction robots to move around job sites and perform required tasks on their own, autonomous navigation capabilities are necessary.

Many prior studies have focused on enabling robots to autonomously move to destination coordinates while considering the obstacles and structures \cite{ref_26, ref_27, ref_28, ref_29}. The robot perceived the environment, using Light Detection and Ranging (LiDAR) \cite{ref_28} or visual information \cite{ref_30, ref_31}, and planned and revised a path to the destination using a planning algorithm such as Rapidly-Exploring Random Tree (RRT) \cite{ref_31} or A* \cite{ref_29}. These studies have enabled construction robots to autonomously move around complex environments while avoiding collisions. However, these approaches are limited in that destination coordinates required for the task execution must be specified in advance \cite{ref_26}. In construction environments with frequent improvisational tasks \cite{ref_15}, task content, locations, and execution order may not be determined in advance, requiring human operators to repeatedly check and reassign destination coordinates.

To reduce the need for manually specifying such coordinates, several studies have derived task-required location coordinates from Building Information Modeling (BIM) \cite{ref_32}. When the user specifies the task-required object in the BIM, the destination coordinates can be set based on recorded location information, without identifying and inputting destination coordinates every time. However, the methods remain constrained by BIM-recorded information. As a result, it remains difficult to handle improvisational tasks that deviate from the original plan and to identify task locations associated with tools, equipment, or material storage that are not explicitly recorded in BIM.

Recent Large Language Model (LLM)- and LMM-based mobile robot navigation studies demonstrate the potential for robots to understand non-predefined tasks given in natural language and identify task-required locations that are not pre-recorded \cite{ref_18, ref_20, ref_21, ref_23}. Leveraging the natural language understanding capabilities, LLMs can interpret operators’ commands that are not explicitly pre-specified in advance \cite{ref_33}. In addition, vision-integrated LMMs can combine environmental perception with robot navigation, enabling the identification of task-relevant locations that are not pre-specified \cite{ref_34}. While promising, LLM- and LMM-based robot navigation is still in its early stages and struggles to address the challenges posed by improvisational tasks on construction sites. We highlight three key challenges.

First, many LLM- or LMM-based navigation studies still face limitations in identifying task-required locations that are not pre-recorded in construction environments. In many prior works, the LLM selected destinations from a set of pre-recorded coordinate candidates based on the user commands \cite{ref_21, ref_22}. A recent study explored using LLMs to interpret BIM data and set a destination based on it, but such approaches still relied on pre-recorded location information \cite{ref_23}. To reduce reliance on pre-recorded information, several studies in robotics have combined LLM reasoning with visual perception to locate target objects that are not pre-recorded \cite{ref_18, ref_19, ref_20, ref_35}. However, the methods are not usually tailored to the characteristics of construction sites. Many prior studies focused on daily-life environments (e.g., residential homes or office rooms) \cite{ref_18, ref_19, ref_35, ref_36} and leveraged the semantic meanings of spaces (e.g., bedroom, living room, or kitchen) as a cue for navigation \cite{ref_18, ref_19, ref_36}. Because such semantic meanings are often difficult to identify from visual information in under-construction sites, it is necessary to further tailor these methods by incorporating construction-specific cues (e.g., construction drawings).

Second, previous studies have limitations in addressing non-predefined contextual or attribute conditions in improvisational tasks in construction. Previous studies that used visual information to find targets were often limited to simple object searches without additional conditions (e.g., “find a bag”) \cite{ref_18, ref_19, ref_35}. However, for improvisational tasks in complex construction environments, task targets often involve attributes (e.g., shape, color, or size) or contextual conditions (e.g., relative locations) to clearly specify the task target. Improvisational task instructions may include conditions such as “check the fire extinguisher without a yellow inspection tag” or “move the box blocking the north stairway.” A few recent studies have investigated enabling robots to identify and navigate to targets defined by attribute conditions through open-vocabulary detection models \cite{ref_37, ref_38}. However, such open vocabulary detection models are limited to distinguishing several simple attribute conditions and still struggle to reason about complex attributes or contextual conditions, as further discussed in our results.

Third, flexibility in the timing of task assignments remains limited. One of the key characteristics of improvisational tasks on construction sites is that they can be assigned and modified at any time. In many LLM and LMM-based navigation approaches, systems operate in a one-directional manner, meaning that a new command can be processed only after the previous one has been completed \cite{ref_18, ref_20, ref_39, ref_40}. Therefore, in many studies, it was difficult to process additional commands during task execution, such as modifying the task instructions, changing the task order, adding new tasks, and canceling the task \cite{ref_18, ref_20, ref_40}.

Thus, LMM-based robot navigation provides fundamental capabilities for understanding non-predefined commands and identifying task-required locations. However, the three limitations outlined hinder its applicability to improvisational tasks on construction sites and must be resolved to enable mobile construction robots to autonomously identify task-required locations and position themselves accordingly.

\section{AI agent with multi-LMM module}
\label{sec:Method}

We proposed an LMM-based agent for a mobile construction robot that can autonomously position itself in response to improvisational tasks involving unknown task-required locations, non-predefined attributes, or contextual conditions, and mid-execution task assignment and modifications. 

\begin{figure*}[!t]
\centering
\includegraphics[width=0.9\textwidth]{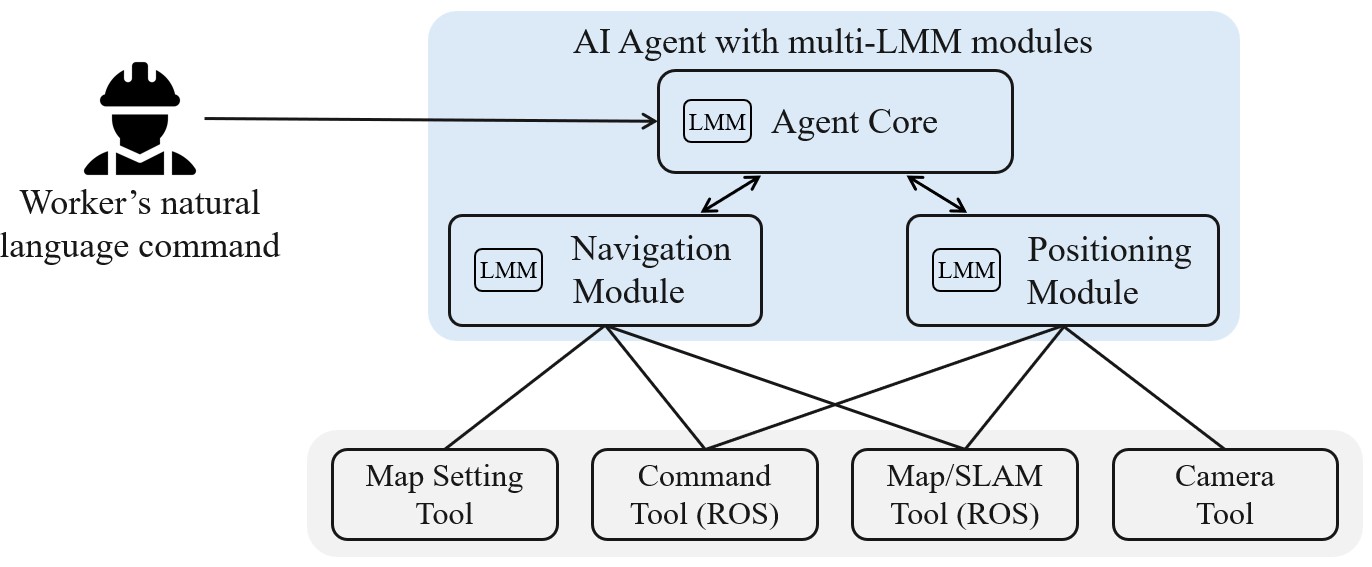}
\caption{Framework of the proposed AI agent with multiple modules}
\label{fig_1}
\end{figure*}

\begin{figure*}[!b]
\centering
\includegraphics[width=\textwidth]{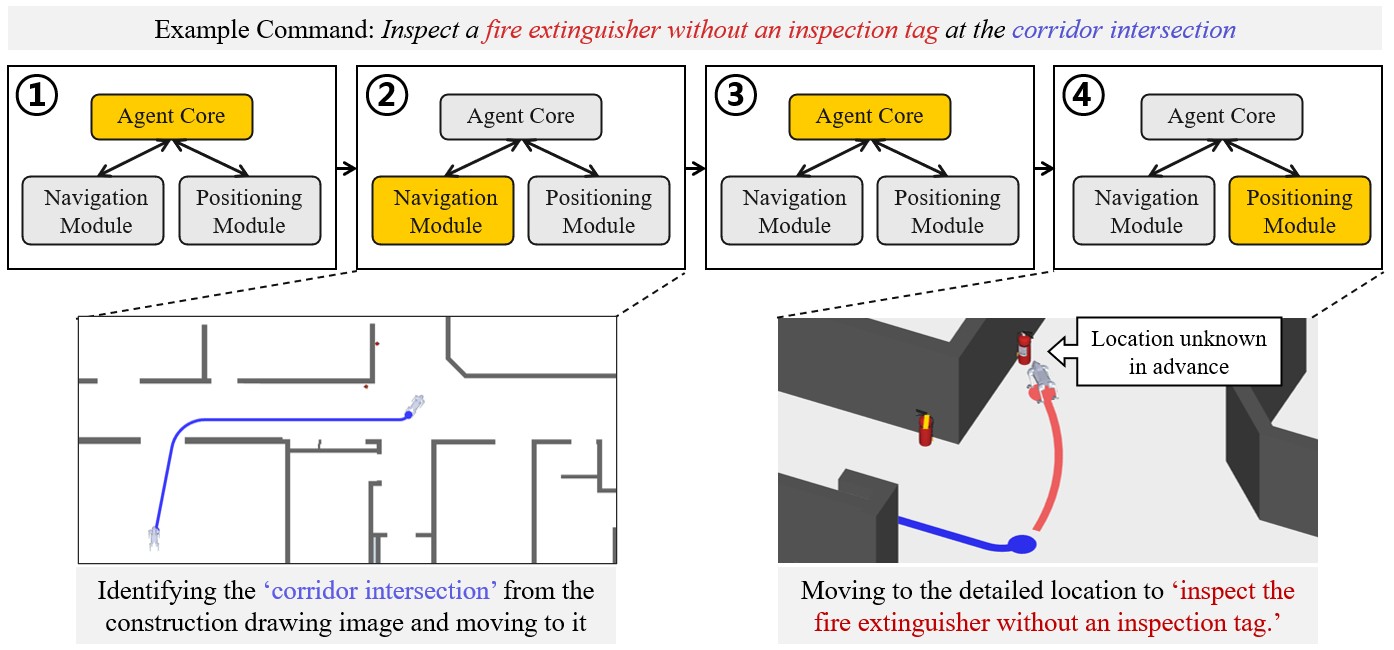}
\caption{Conceptual overview of the proposed agent’s operation}
\label{fig_2}
\end{figure*}

We construct an agent composed of three LMM reasoning modules. Figure \ref{fig_1} shows the framework of the agent we propose. We adopted a framework that integrates characteristics of both the LMM workflow and the LMM-based agent, leveraging their complementary strengths while mitigating their respective limitations. An LMM workflow follows a predefined sequence of LMM questioning in order. A workflow provides greater control over the output, ensuring the final output appears in the intended form, but it lacks flexibility when encountering new situations or commands \cite{ref_41}. In contrast, an AI agent iteratively interprets the environment based on the given objective, identifies the necessary tools or methods, performs reasoning, and decides when the task has been completed on its own, enabling more flexible task execution \cite{ref_41, ref_42}. However, compared to a workflow, AI agents have less control over the final output \cite{ref_41}. In our framework, each module adopts characteristics of the workflow to ensure it performs the intended roles, while the overall framework is designed to function as an agent, providing the flexibility needed to handle improvisational tasks. Furthermore, by separating the system into three modules, we distribute complex reasoning tasks across them, reducing the burden on any single component. The modules operate in parallel, enabling the agent to respond flexibly to mid-execution commands that may arise during task execution. While the current framework employs three modules, it can be extended by placing additional LMM modules under the agent core to incorporate new functionalities as needed.

Figure \ref{fig_2} conceptually illustrates how the proposed agent operates. Our approach uses the location information that can be acquired from the construction drawing as much as possible, while supplementing detailed and non-static elements that are not recorded on the drawing with visual information. First, the agent core receives a natural language command from the user and breaks it down into positional information that can be known from the construction drawing and detailed task instructions. Next, the navigation module interprets the construction drawings provided as a Portable Network Graphics (PNG) image, identifies the approximate zone-level location (e.g., rooms, corridors, elevators, and stairways) for task execution, and moves the robot to that location. Once navigation is complete, the agent core passes the task information to the positioning module. The positioning module then uses visual information from the robot’s camera to move the robot to the detailed task-required position. After that, the positioning module reports the task completion status to the agent core, and the agent core proceeds to execute the next pending task.

\begin{figure*}[!t]
\centering
\includegraphics[width=0.9\textwidth]{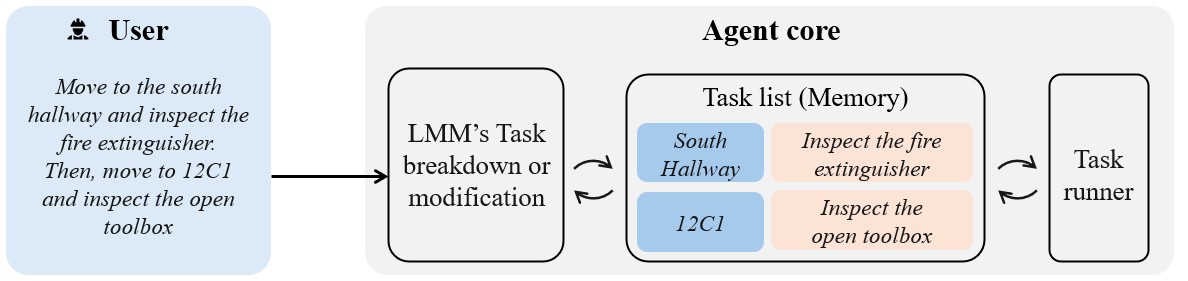}
\caption{Three components of the agent core and their processing}
\label{fig_3}
\end{figure*}

\subsection{Agent core}

\begin{figure}[!htb]
\centering
\includegraphics[width=\columnwidth]{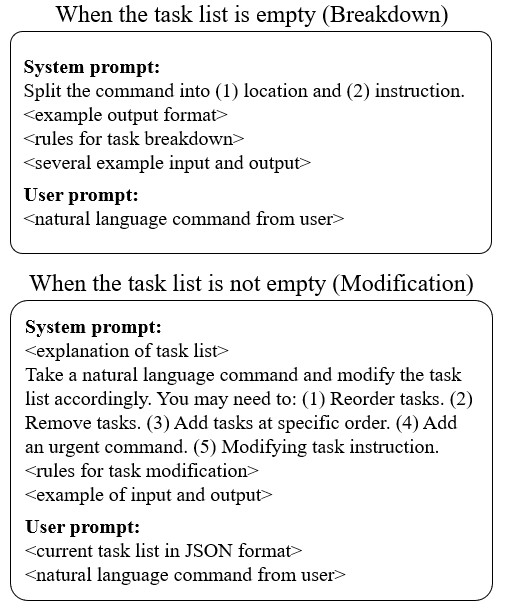}
\caption{Prompts for agent core}
\label{fig_4}
\end{figure}

The agent core consists of three components: the LMM’s reasoning, the task memory, and the runner. Figure \ref{fig_3} visualizes the three components. The LMM’s reasoning generates or modifies the task list in response to natural-language commands. When generating this list, the command is also broken down into an approximate zone-level location and a detailed task instruction. Figure \ref{fig_4} shows part of the prompt provided to the LMM. LMM reasoning is called only when a new command is provided by the user. When the task list is empty, the LMM performs the initial breakdown; when tasks remain in the list, it proceeds with modification.

The task runner operates in an independent asynchronous loop that runs in parallel with the LMM reasoning. The task runner pops tasks from the top of the list and sends the zone-level location information to the navigation module and the task instruction to the positioning module. For the task runner, timing is taken into account. It waits for the navigation result, proceeds with positioning only if navigation succeeds, and marks the entire task as failed if navigation fails. In the task modification, when the currently running task needs to be changed, the task runner sends a stop signal to both subordinate modules and then reissues the newly modified task instructions.

\subsection{Navigation module}

The navigation module is responsible for navigation toward an approximate zone-level location based on the known information provided in the construction drawing. The navigation module operates with two preloaded inputs: a construction drawing in image format with zone labels (text tags indicating zones such as rooms, corridors, elevators, and stairways), and a Robot Operating System (ROS) binary grid map. First, it is common for construction sites to have at least basic 2D drawings, making the use of construction drawings reasonable. Second, in this study, we assume that the robot has an initial ROS binary grid map showing the occupancy of space. Therefore, the initial setup of this binary map is required once the robot is first deployed in a new environment. However, since many studies have already investigated autonomous initial exploration methods for generating binary occupancy maps, this step could also be integrated into future versions of our system \cite{ref_43, ref_44}.

\begin{figure*}[!t]
\centering
\includegraphics[width=\textwidth]{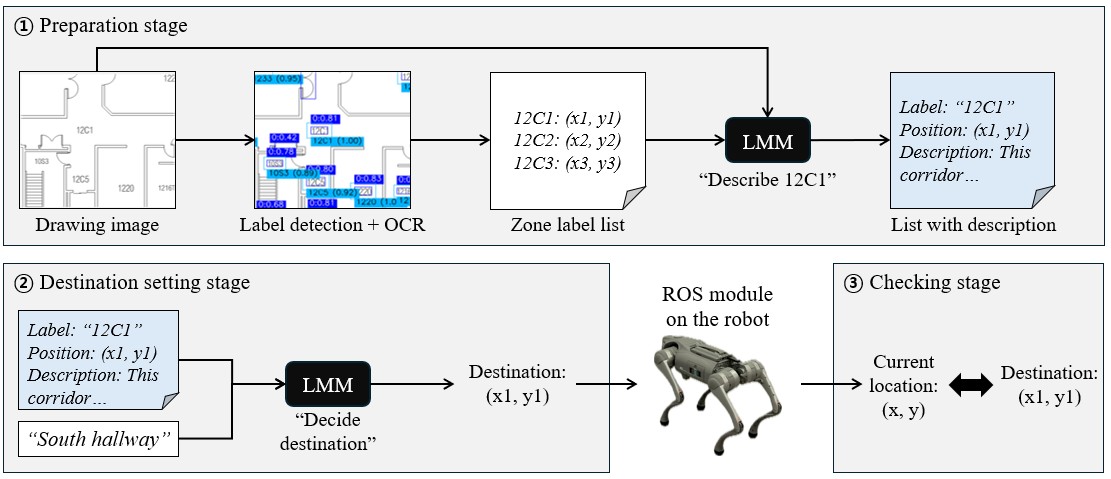}
\caption{Three-stage process of the navigation module}
\label{fig_5}
\end{figure*}

The operation of the navigation module can be divided into three stages: a preparation stage that analyzes the construction drawing before task execution begins, a destination setting stage that responds to commands from the agent core once the task has started, and a navigation checking stage where the status of movement is checked to determine whether the navigation is complete. Figure \ref{fig_5} illustrates this process.

The objective of the first preparation stage is to create JSON files that contain the zone label name, the position of each zone, and the textual description associated with that label’s location. The module first detects the zone labels on the construction drawing image and extracts their text information using an Optical Character Recognition (OCR) model. A pretrained YOLO model and the EasyOCR package were used for this process \cite{ref_45, ref_46, ref_47}. Detected label bounding boxes are then converted into real-world coordinates (in meters) through a transformation between the construction drawing and the actual map. This process produces a JSON file that stores each label name along with its corresponding location.

\begin{figure}[!htb]
\centering
\includegraphics[width=\columnwidth]{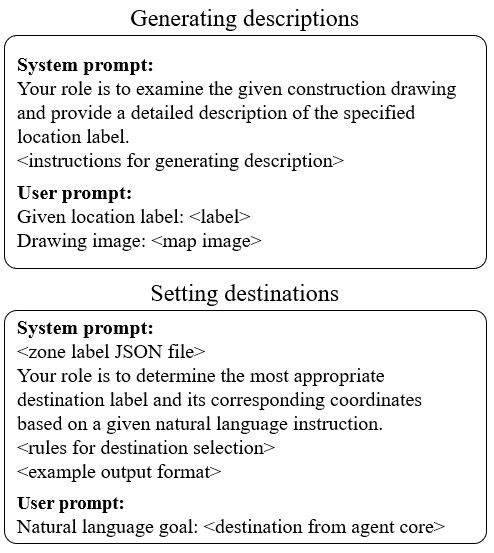}
\caption{Prompts for the navigation module}
\label{fig_6}
\end{figure}

Next, LMM generates a description for each label location. The purpose of generating these descriptions is to ensure flexibility in determining the intended location, even when the user’s command does not exactly match the zone label name but instead refers to it semantically (e.g., south hallway, the hallway in front of the elevator, and hallway intersection). Generating text descriptions prior to task execution improves accuracy, as LMMs can more effectively match instructions to textual descriptions than directly identify locations such as “south hallway” from drawing images \cite{ref_48}. Figure \ref{fig_6} shows the prompt for generating descriptions. In the prompt, we use the term ‘location label’ to emphasize that the labels refer specifically to spatial locations.

\begin{figure*}[!b]
\centering
\includegraphics[width=\textwidth]{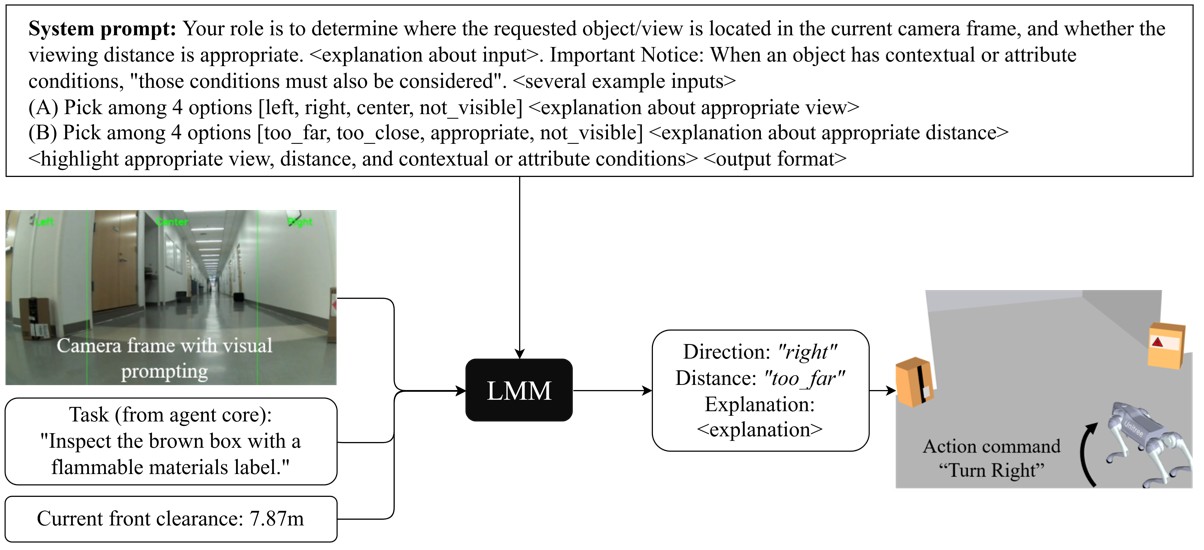}
\caption{Process for positioning module}
\label{fig_7}
\end{figure*}

In the destination setting stage, based on the generated JSON file, when a command is provided by the agent core, the module determines the destination coordinates and sends them to the robot’s ROS. When the user’s command specifies an exact label name, the destination is determined directly without LMM reasoning. When the instruction is given in a semantic form, such as “south hallway” or “the hallway in front of the elevator”, the LMM reviews the descriptions in the JSON file and selects the most appropriate destination. In addition, when given a new location, expressions such as “center between 12C1 and 12C2” are also supported, allowing the destination to be specified through LMM’s simple calculation based on the positions of the referenced labels. The prompt is briefly introduced in Figure \ref{fig_6}. The final selected coordinate is transmitted to the robot’s onboard ROS system, which uses SLAM and the Navigation2 packages to autonomously move to the designated coordinate \cite{ref_49}. 

Finally, in the navigation checking stage, the robot’s position output from the ROS is received to determine whether the navigation has been successfully completed. Once the navigation is completed, the result is reported to the agent core. If the navigation stops abnormally (e.g., when the robot becomes too close to an obstacle), a predefined LiDAR-based recovery command is issued to move the robot away from the obstacle, after which the navigation process is restarted. 

\subsection{Positioning module}

After reaching the approximate zone-level location, the robot uses visual information to identify and move to a more precise task-required location. By leveraging visual information, the positioning module can identify task-required locations that are not specified in the construction drawing. Rather than employing an additional object detection model, our module allows the LMM to directly interpret the camera frame and infer the direction and distance to the task-required location. This design enables the positioning module to process attributes and contextual conditions during task-required location identification. 

Figure \ref{fig_7} visualizes the operation of the positioning module. The LMM is provided with the three inputs: (1) the current camera image; (2) the task instruction (from the agent core); and (3) the forward clearance. Based on the inputs, the LMM repeatedly outputs the direction and distance to the task target specified in the task instruction. For direction, the LMM chooses one of four options: left, right, center, or not visible. To assist the LMM in determining direction, we applied visual prompting by dividing the frame into regions annotated as left, center, and right \cite{ref_50}. For distance, LMM selects from too far, too close, appropriate, or not visible.

To assist in distance adjustment, the forward clearance distance is provided as an input based on the LiDAR data. However, this value represents only the forward clearance and not the distance to the target, so it must be interpreted in conjunction with visual information. The appropriate distance to the target is user-adjustable by prompt. In addition to questions about direction and distance, the LMM is required to provide reasoning for its decision, as generating such explanations can improve the accuracy of its responses \cite{ref_51}. During the questioning, the LMM is instructed to carefully consider any relevant attributes or contextual conditions included in the instruction.

\begin{figure*}[!b]
\centering
\includegraphics[
    width=0.95\textwidth,
    keepaspectratio
]{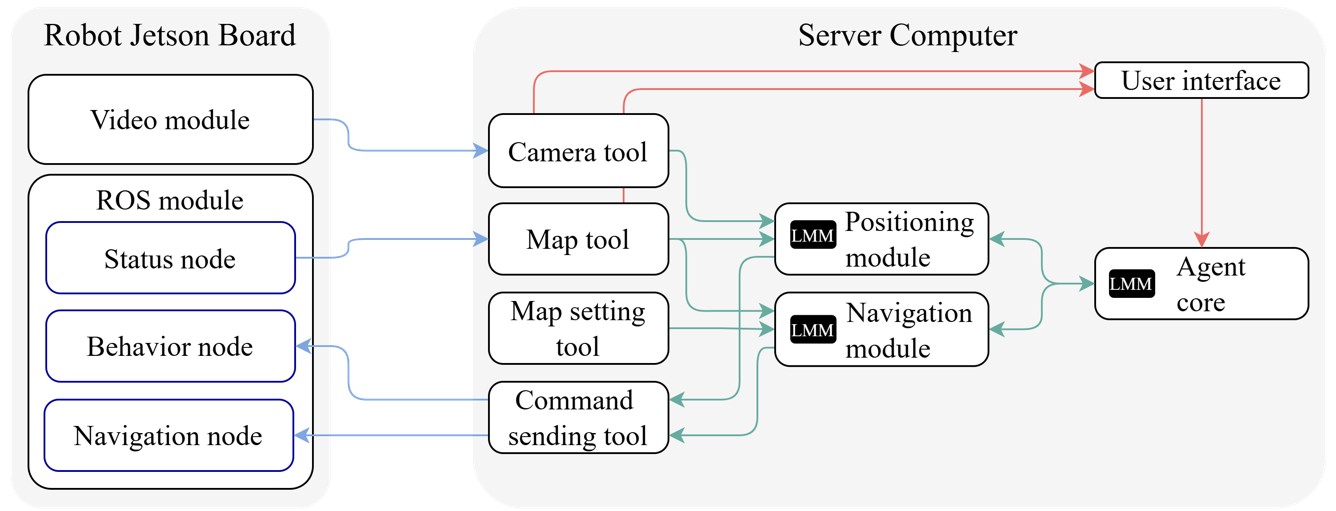}
\caption{Implemented system architecture, illustrating interactions among modules and tools}
\label{fig_8}
\end{figure*}

Short actions are carried out according to the LMM’s decisions. The available actions include move forward (25cm), move backward (25cm), turn left (15º), turn right (15º), and look around (40º). The commands are designed to prioritize aligning the robot’s orientation with the approximate direction of the target. For example, if the LMM’s decision is right, indicating that the target is located to the robot’s right, a turn right command is issued. Once the direction falls within the center range, the module adjusts the distance accordingly. If the distance is classified as too far, the system executes a move forward action. If the LMM outputs are not visible, a look-around command is issued, which rotates the robot through a wider angle to search for other directions. By repeating this process, the module considers the task complete once the direction is classified as center and the distance as appropriate, and then sends the result to the agent core.

\section{Implementation of the proposed agent framework with a quadruped robot}

We implemented the system, integrating the proposed agent framework with a quadruped robot, the Unitree Go2 [24]. For the agent framework to be integrated with a real robot, the decision-making processes of each LMM module, the robot’s localization and camera data acquisition, and the ROS-based robot control must be organically integrated and exchange information in real-time. To reduce the computational load on the robot’s onboard computer, processes requiring low latency (within 100ms) are executed on the robot’s Jetson board, whereas other components that allow a few seconds of latency are executed on an external computer. Figure \ref{fig_8} visualizes the implemented system architecture.

We designed a multi-Docker-container backend structure that allows each module and its associated tools to operate independently and in parallel \cite{ref_25}. This Docker-based architecture prevents dependency conflicts among modules and provides high scalability for adding new tools \cite{ref_25, ref_52}. Communication between containers within the same machine was implemented using the Google Remote Procedure Call (gRPC) protocol to enhance transmission reliability \cite{ref_53}, which has also been widely used for inter-container communication in previous studies. Communication between the robot and the server was carried out via Transmission Control Protocol (TCP), which provides low latency and minimal overhead. Communication with the user interface was carried out over HyperText Transfer Protocol (HTTP), which is widely supported by web-based user interface frameworks.

\subsection{On-robot implementation}

Two Docker containers are deployed on the robot’s Jetson board (Jetson Orion Nano). The first container is the ROS module container. It utilizes the Unitree ROS software development kit to interface with the robot’s hardware, while communication is managed through the Data Distribution Service (DDS) framework of ROS 2. We used ROS packages that handle real-time LiDAR point accumulation, map matching using slam-toolbox, and navigation using the Navigation 2 package \cite{ref_49, ref_54, ref_55}. 

The ROS module receives two types of commands from the server. Here, we referred to these two types of commands as the “navigation command” and the “behavior command,” respectively. First, the “navigation command”, which is set as the target point for path planning using the A* algorithm of the Navigation 2 package \cite{ref_49, ref_54}. It is sent from the Figure \ref{fig_8} command-sending tool to the ROS module’s navigation node. Second, “behavior command” is the single-action motion command, which includes actions such as “move forward”, “move backward”, “turn right”, “turn left”, “omni-directional move right”, “omni-directional move left”, and “stop”. The status node transmits the robot’s status data to the server. Robot’s status data includes the robot’s current map coordinates (i.e., x, y coordinates in meters) and orientation obtained from the Navigation 2 package, and the rough clearance distances to the front, rear, left, and right, calculated from LiDAR data. The second container handles video processing. It receives multicast video streams using GStreamer, encodes them, and transmits the encoded video to the server \cite{ref_56}.

\begin{table*}[!b]
\centering
\caption{Improvisational task challenges in each test}
\label{tab:1}

\renewcommand{\arraystretch}{1.1}
\begin{tabularx}{\textwidth}{cYYY}
\toprule
Test &
Need to identify and move to the task-required location &
Need to consider attribute and contextual conditions &
Need to address mid-execution task assignments and adjustments \\
\midrule
A & Yes & No  & No  \\
B & Yes & Yes & No  \\
C & Yes & Yes & Yes \\
\bottomrule
\end{tabularx}

\end{table*}

\subsection{On-server implementation}

The server consists of eight containers in total. Three of them are LMM modules (i.e., the agent core, the navigation module, and the positioning module). We used GPT-4o as the LMM in these modules. Four containers function as tools connected to these modules, and the last container hosts the user interface for inputting commands. The four tools perform the following roles. First, the map setting tool includes a text detection model and an OCR model, and is invoked by the navigation module to detect and read text labels from the construction drawing in image format. Second, the map tool receives the robot’s real-time map position, orientation, and front, back, left, and right clearance information. This information is forwarded to the navigation module, which uses it to assess navigation progress. Third, the command sending tool receives navigation commands and behavior commands from the navigation or positioning modules and forwards them to the robot. When forwarding a command, it performs rule-based validation to ensure the command is in a valid format and to prevent unintended behavior. Fourth, the camera tool continuously receives video frames from the robot’s streaming module and stores the latest frame. For testing, a simple web-based user interface was implemented. The interface supports natural-language command input from the user. Although not essential for the operation, the robot’s location and camera feeds were displayed on the user interface through the map and camera tools to monitor and record robot status during the tests.

\section{Testing}
\label{sec:Implementation}

We tested the performance of the proposed agent in identifying task-required locations and positioning itself to them in response to improvisational tasks. In particular, we aim to evaluate the agent’s ability to handle three key challenges in positioning for improvisational tasks: (1) unknown task-required locations; (2) attribute- and contextual conditions; and (3) mid-execution task assignments and adjustments. Three tests are designed to progressively incorporate these three key challenges. The challenges addressed in each test are summarized in Table \ref{tab:1}. Test A evaluates the first challenge by assessing whether the proposed agent can interpret an improvisational task command and identify and move to a task-required location whose position is unknown. Test B builds on the setting of Test A by additionally assessing whether the agent can identify and move to the task-required location by considering attributes or contextual information contained in the command. Test C further extends the setting of Test B by evaluating whether the agent can address commands that are added or modified during task execution. Test C represents the most comprehensive case, requiring the agent to address all three challenges.

\subsection{Sessions and tasks}

In each of the three tests, the agent executes 15 test sessions. A session is defined as a continuous sequence of operations performed without manual control. Within a single session, multiple improvisational tasks and multiple natural language commands, including mid-execution commands, may be given. We evaluate how many sessions are successfully completed in each test. A session is considered successful if the following two conditions are satisfied: (1) all improvisational tasks in the session are successfully completed in terms of positioning; and (2) the tasks are performed in the correct order, including any modifications such as reordering, additions, or removals given by mid-execution commands.

Each session contains one to three tasks, and all tasks must be successfully completed for the session to be considered successful. A task is defined as a unit in which the agent identifies the location required to accomplish a given instruction and moves to that location. We designed the test tasks as improvisational inspection tasks. Our focus is not on the execution of the task itself (e.g., carrying, welding, and assembling), but on the agent’s ability to identify and reach the task-required location. Improvisational inspection tasks represent the minimal form of tasks that allow us to evaluate positioning capabilities without being affected by manipulation or other execution processes. The positioning performance evaluated through these inspection tasks can be extended to other types of improvisational tasks. For example, an improvisational task like “Carry these bricks and place them on the brick pile in zone 12C4” fundamentally relies on the robot’s ability to identify and position itself at the brick pile in 12C4.  

A task is considered successful if the following three conditions are satisfied. First, the robot identifies and faces the task-required target (i.e., the inspection target). Second, the robot approaches the target within one meter. In prior studies, particularly those focusing on object searching where the target location is not known in advance, this threshold is often not explicitly defined. When it is specified, relatively large allowable distances, such as 1.5 meters or even 3 meters, are also permitted \cite{ref_57, ref_58}. However, our focus is on positioning rather than object searching, and thus the threshold is set to one meter. Based on this threshold value, the agent’s positioning module was prompted to maintain an approximate distance of 0.5 to 0.8 meters as the most appropriate range. Distances closer than 0.5 meters are avoided because, if the robot approaches the task target too closely, the target cannot be fully observed due to the Go2 camera’s field-of-view configuration. Third, the agent itself judges the task as successful. This third criterion reflects that the robot operates as an agent, and thus its ability to judge task success was an important consideration.

Table \ref{tab:2} shows the number of tasks and sessions for each test. We evaluated performance using session and task success rates. The metric Success Rate (SR) has been widely used for measuring the performance of the autonomous task execution of the robots \cite{ref_20, ref_57}. The task success rate is defined as the ratio of successfully completed tasks to the total number of tasks in each test. The session success rate is defined as the ratio of successfully completed sessions to the total of 15 sessions in each test. 

\begin{table}[!htbp]
\centering
\caption{Number of sessions and tasks in each test}
\label{tab:2}
\renewcommand{\arraystretch}{1.1}

\begin{tabularx}{\columnwidth}{cYY}
\toprule
Test &
No. of sessions &
No. of tasks \\
\midrule
A & 15 & 35   \\
B & 15 & 35  \\
C & 15 & 33 \\
\midrule
Total & 45 & 103 \\
\bottomrule
\end{tabularx}
\end{table}

\subsection{Commands and test environment settings for each test}

\begin{figure}[!htb]
\centering
\includegraphics[width=\columnwidth]{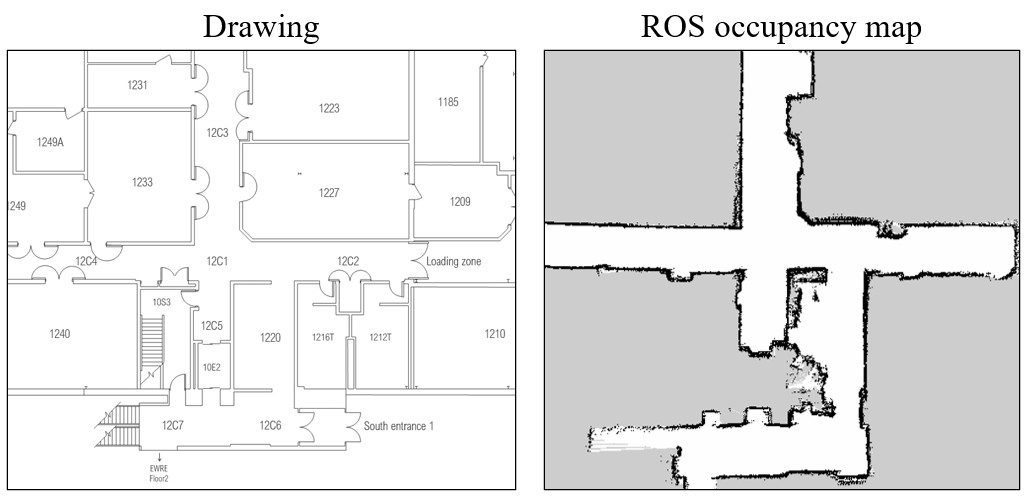}
\caption{Prompts for the navigation module}
\label{fig_9}
\end{figure}

All three tests were conducted in an indoor office building simulating an indoor construction site, as shown in Figure \ref{fig_9}. The figure on the left shows the construction drawing used in the experiments, while the figure on the right shows the ROS map. We chose an indoor environment where diverse zone-level locations can be defined, and task-relevant visual elements can be easily controlled, making it suitable for generating diverse improvisational tasks.

The setting of task targets, the information contained in the commands, and the timing at which the commands are given differ across the tests according to their objectives. Table \ref{tab:3} presents example images illustrating the test target settings and the corresponding commands for each test. The underlined parts in Table \ref{tab:3} indicate the task currently being executed in each image.

\newcolumntype{C}[1]{>{\centering\arraybackslash}m{#1}}
\newcolumntype{L}[1]{>{\raggedright\arraybackslash}m{#1}}

\begin{table*}[!t]
\centering
\caption{Example task target setup and example commands for Test A, B, and C}
\label{tab:3}

\begin{tabular}{C{0.08\textwidth} C{0.4\textwidth} L{0.4\textwidth}}
\toprule
\multicolumn{1}{c}{Test} &
\multicolumn{1}{c}{Example task target setup} &
\multicolumn{1}{c}{Example command} \\
\midrule

A &
\parbox[c][3.4cm][c]{\linewidth}{\centering
\includegraphics[width=0.95\linewidth, trim={0 5 0 5}, clip]{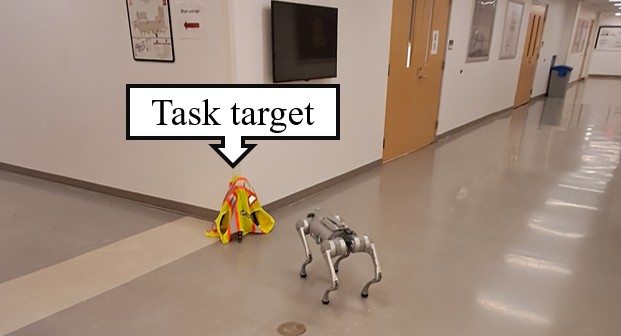}
} &
\parbox[c][3.4cm][c]{\linewidth}{\raggedright
Move to the south hallway and inspect the fire extinguisher. \uline{\textbf{Then move to the hallway}} \uline{\textbf{intersection to inspect the safety vest.}} Then move to 1220 and inspect the brown box.
} \\

\midrule

B &
\parbox[c][3.4cm][c]{\linewidth}{\centering
\includegraphics[width=0.95\linewidth, trim={0 5 0 5}, clip]{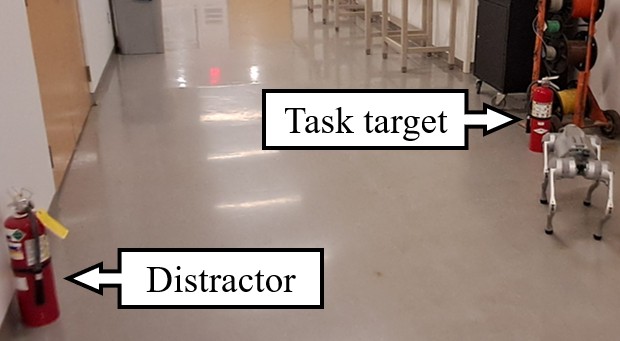}
} &
\parbox[c][3.4cm][c]{\linewidth}{\raggedright
Move to 12C5 and inspect the open toolbox. \uline{\textbf{Then move to 12C3 to inspect the fire}} \uline{\textbf{extinguisher without a yellow tag.}}
} \\

\midrule

C &
Both settings with and without distractors are included. &
(Mid-execution command) Actually, inspect the hand-pulled cart carrying fire extinguishers instead of the one carrying the brown boxes.
\\

\bottomrule
\end{tabular}
\end{table*}

In test A, the agent is required to identify and move to the task-required location. In each session, inspection tasks involve objects as the task target that are commonly found on construction sites but are usually not specified in construction drawings (e.g., toolbox, hand-pulled cart, tripod, hard hat). The locations of these task targets are randomly assigned, and their exact locations are not known to the agent. Examples of commands for Test A are shown in Table \ref{tab:3}. Each command includes an approximate zone-level location where the task is to be performed, along with detailed task instructions. Among 15 sessions, ten sessions include two tasks provided as a single natural language command at the beginning of the session, while the remaining five sessions include three tasks. In addition, in five sessions, at least one zone-level location is described using descriptive terms (e.g., “south hallway”) rather than explicit label names defined in the drawings. 

Test B builds upon the setting of Test A by requiring the agent to consider contextual and attribute conditions when identifying task locations. Test B follows a similar test target setup to Test A. However, in Test B, distractors are placed in close proximity to the target objects. Here, distractors are intentionally placed objects that can lead the agent to an incorrect task location if the specified conditions are not properly considered. Distractors were placed for all 35 tasks across the 15 sessions. The agent must avoid the distractors and position itself at the location required by the task instructions. Table \ref{tab:3} illustrates an example task target setup in which distractors are present. As in Test A, all tasks were given as a single command at the beginning of each session.

Test C extends the setting of Test B. The main difference lies in the way the commands are given. In test C, initial commands are provided at the beginning, and one or two mid-execution commands are issued during execution to add new improvisational tasks, modify the content or order of ongoing tasks, or cancel tasks. The agent must reflect these updates in real time and execute the tasks as intended for the session to be considered successful. Among the 15 sessions, 3 involve task additions, 3 involve cancellations, 3 involve order changes, 3 involve modifications to task instructions, and the remaining 3 involve a combination of these mid-execution command types. Table \ref{tab:3} provides an example of mid-execution commands given in Test C. In Test C, tasks include both cases with and without contextual conditions, and the agent is required to consider contextual conditions when necessary based on the given command.

\subsection{Comparative test on identifying task target considering attributes and contextual conditions}

Although no prior work exactly matches the scope of this study, a limited number of approaches have explored robot navigation toward targets with attribute conditions (e.g., finding “a tall white trash can”) \cite{ref_37, ref_38}. In these approaches, open-vocabulary detection models were integrated to distinguish task targets from distractors, using detection confidence scores as the primary cue. However, while open-vocabulary detection model-based approaches may distinguish relatively simple attribute conditions, they have limited capability in reasoning over complex contextual or relational conditions.

We compared the target identification capability of the proposed agent with open-vocabulary detection model-based approaches under the same task target and distractor configurations. A total of 15 tasks were provided in the Test B setting. A comparison was conducted using state-of-the-art, widely used open-vocabulary models, Grounding DINO and YOLOv8x-world \cite{ref_59, ref_60}. Here, we focus on comparing the ability to identify target objects under attribute and contextual conditions, using a simplified setup to isolate the target identification capability from the overall positioning process. We assumed navigation to the zone-level location was completed. For open-vocalbuary models, the robot captured images in all directions at the zone-level location and selected the object with the highest confidence score as the target, which was then evaluated to determine whether it corresponded to the commanded target or a distractor. Here, success was defined solely by whether the robot correctly selected the target object.

\section{Results}

Tests were conducted under the settings described for Tests A, B, and C, and the task success rate and session success rate were computed for each. Table \ref{tab:4} reports on the task and session success rates for each test. Across three tests, a task success rate of 92.2\% and a session success rate of 82.2\% were achieved.  A demonstration of Unitree Go2 with the proposed agent performing the tests can be seen in following YouTube link (https://youtu.be/jh2QuK7ue6E).

\begin{table*}[!t]
\centering
\caption{Measured success rates in each test}
\label{tab:4}

\renewcommand{\arraystretch}{1.1}
\begin{tabularx}{\textwidth}{c|YY|YY|YY}
\toprule
Test & Success task & Failed task & Success session & Failed session & Task success rate & Session success rate \\
\midrule
A & 33 & 2  & 13 & 2 & 94.3\% & 86.7\%  \\
B & 31 & 4 & 11 & 4 & 88.6\% & 73.3\%  \\
C & 31 & 2 & 13 & 2 & 93.9\% & 86.7\% \\
\midrule
Total & 95 & 8 & 37 & 8 & 92.2\% & 82.2\% \\
\bottomrule
\end{tabularx}
\end{table*}

\begin{table*}[!t]
\centering
\caption{Comparison of attribute and contextual condition handling in Test B setting}
\label{tab:5}

\renewcommand{\arraystretch}{1.1}
\begin{tabularx}{\textwidth}{cYYY}
\toprule
Target identification method & Success Task & Failed Task & Success Rate \\
\midrule
Open-vocabulary model (Grounding DINO) & 4 & 11  & 26.7\% \\
Open-vocabulary model (Grounding DINO) & 5 & 10  & 33.3\% \\
Ours & 13 & 2  & 86.7\% \\
\bottomrule
\end{tabularx}
\end{table*}

\subsection{Results for Test A}

Test A has the most basic setting among the three tests, requiring the agent to find the task target with an unknown location by integratively using drawing and visual information. The task success rate was 94.3\%, and the session success rate was 86.7\%, achieving the highest performance among the three tests. 

While direct comparison is difficult because no existing studies share the exact scope and objectives, the success rate of over 90\% suggests competitive performance relative to prior studies addressing autonomous robot navigation toward targets with unknown locations \cite{ref_18, ref_20, ref_37}. In particular, rather than simply searching for the target, a relatively conservative success criterion was adopted—requiring the robot to reduce the distance to within 1 m, face the target, and autonomously determine task completion—and the agent still demonstrated high task success rates of more than 90\% \cite{ref_57}. Furthermore, a sequence success rate of over 85\% was achieved even under a stringent criterion that requires all tasks in a sequence to be successfully executed in the intended order, including those involving mid-execution commands.

\subsection{Results for Test B}

In Test B, the agent is additionally required to identify and move to the task target by considering attribute and contextual conditions specified in the improvisational task instruction. The task success rate was 88.6\%, and the session success rate was 73.3\%. Although the success rate was lower than in Test A, it still achieved a success rate in the upper 80\% range, demonstrating the ability to handle attribute and contextual conditions in the improvisational tasks in construction.

\begin{figure}[!htb]
\centering
\includegraphics[width=\columnwidth]{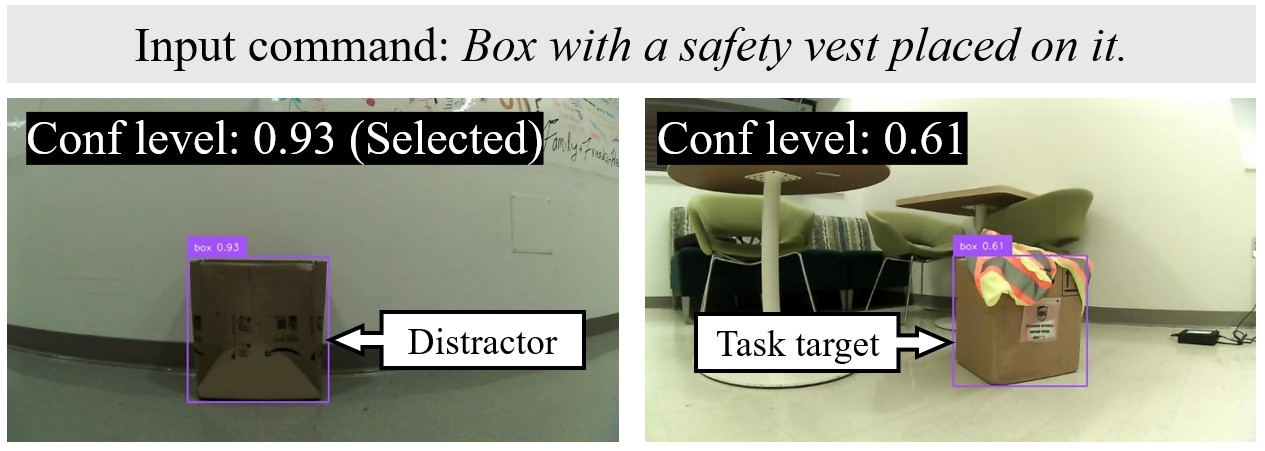}
\caption{Example of a failure case in contextual condition handling using Grounding DINO}
\label{fig_10}
\end{figure}

The comparative results under the Test B setting are presented in Table \ref{tab:5}. Out of 15 tasks, our agent achieved an 86.7\% success rate in correctly identifying the target object while avoiding distractors. However, Grounding DINO and YOLOv8x-world achieved only 26.7\% and 33.3\%, respectively. Figure \ref{fig_10} shows an example of a failure case in which Grounding DINO confuses the target with a distractor. Compared to previous open-vocabulary-based approaches that rely on detection confidence scores for target selection, our agent achieved higher performance in handling attribute and contextual conditions when following contextual and attribute conditions in improvisational task instructions.

\subsection{Results for Test C}

In Test C, the agent must be able to incorporate additional or modified tasks at any point during execution. The task success rate reached 93.9\%, while the sequence success rate was 86.7\%. Despite the presence of mid-execution commands, the performance was similar to that of Test A. In addition, two failed sessions in Test C were not attributed to the incorporation of additional improvisational commands, but rather to failure in distinguishing distractors and failure to approach the target within 1 m. This indicates that real-time acceptance of additional or modified task commands was achieved without failure in all 15 sessions.

\begin{figure*}[!t]
\centering
\includegraphics[width=\textwidth]{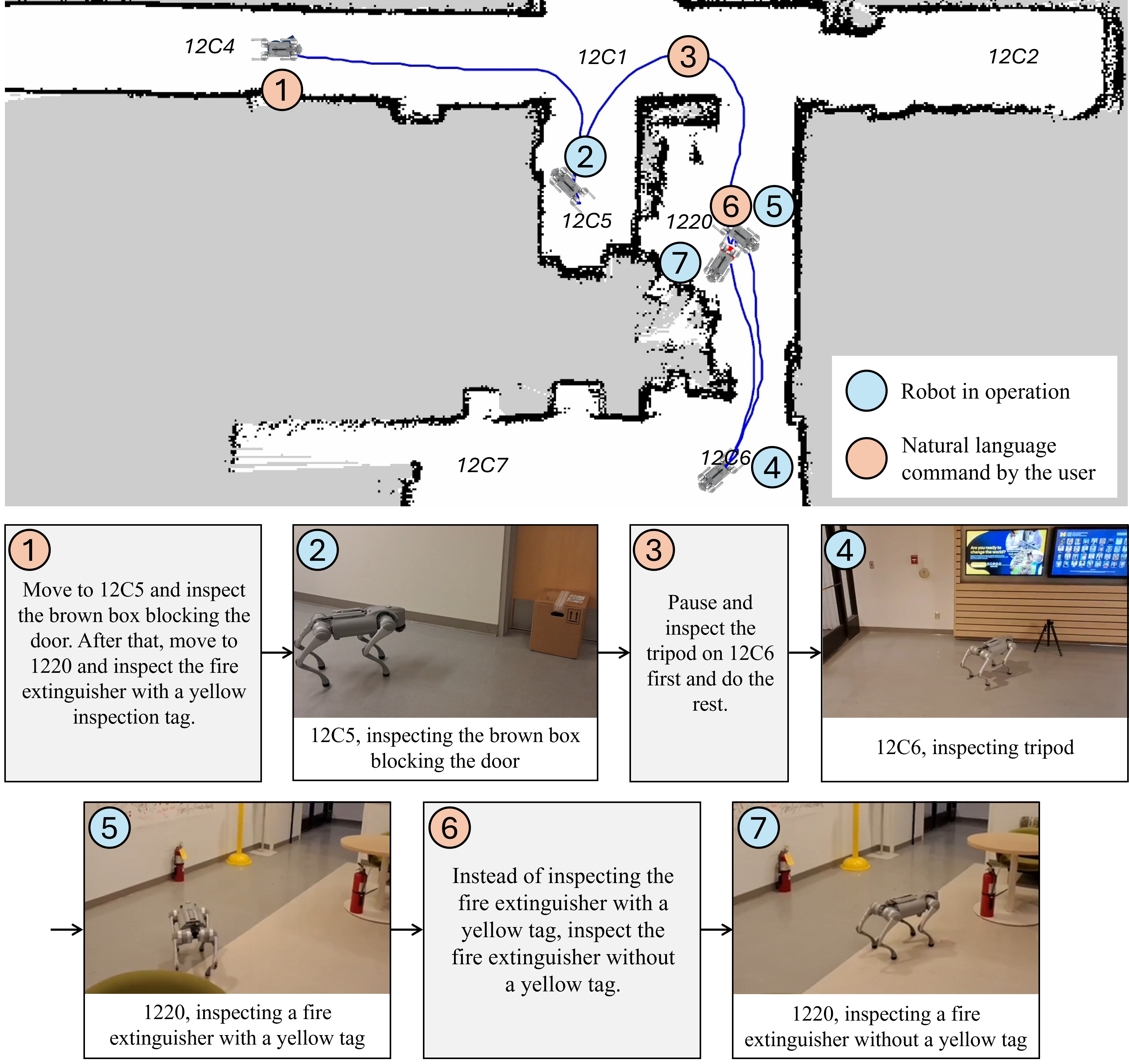}
\caption{Example session on Test C}
\label{fig_11}
\end{figure*}

Figure \ref{fig_11} visualizes a representative case of Test C. Two inspection tasks (i.e., inspect the brown box blocking the door and inspect the fire extinguisher with a yellow inspection task) are given at the start. During the execution, an additional two mid-execution commands are issued. After completing the first task, a higher-priority task (i.e., inspect the tripod) is added while the robot is moving to execute the second task, prompting it to perform the newly added task first. In addition, after completing the tripod inspection, another improvisational command is provided while executing the remaining task. The task instruction is modified to select a fire extinguisher without a yellow tag instead of one with a yellow tag. The robot successfully updates its positioning according to the revised condition. Since this is an example of a single session, the entire process was conducted autonomously without any manual control other than natural language commands.

\section{Discussions}

\begin{table*}[!t]
\centering
\caption{Task failure causes and responsible modules}
\label{tab:6}

\renewcommand{\arraystretch}{1.1}

\begin{tabular}{C{0.18\textwidth} L{0.66\textwidth} C{0.08\textwidth}}
\toprule
Failed Module & \multicolumn{1}{c}{Reason} & Frequency \\
\midrule
Navigation & Wrong zone label selected & 1 \\
\midrule
\multirow{4}{*}{Positioning}
& Failed to find the target (searching until max iterations) & 1 \\
& Positioning to other objects, including distractors & 2 \\
& Final distance was longer than 1 m & 3 \\
& The agent did not judge success as success & 1 \\
\midrule
Total & & 8 \\
\bottomrule
\end{tabular}
\end{table*}

\begin{table*}[!b]
\centering
\caption{Average travel distance and time per session and task}
\label{tab:7}

\renewcommand{\arraystretch}{1.1}
\begin{tabularx}{\textwidth}{c|YY|YY}
\toprule
 & Travel distance per session (m) & Travel distance per task (m) & Time per session (sec) & Time per task (sec) \\
\midrule
Successed sessions & 30.2 & 13.6  & 248 & 111  \\
Failed sessions & 32.6 & 13.9  & 425 & 175  \\
\midrule
All sessions & 30.6 & 13.7 & 280 & 124 \\
\bottomrule
\end{tabularx}
\end{table*}

The proposed agent is capable of autonomously identifying and positioning to the task-required location in response to improvisational tasks. In particular, the success rates measured across the three tests indicate that the three targeted challenges were effectively addressed. The agent can identify and move to a task-required position whose exact locations are not predefined, while considering contextual and attribute-based conditions of the task target. Furthermore, even when improvisational commands are introduced at any point during task execution, the agent can incorporate them into its positioning plan.

Among the three tests, the most important results are from Test C. Test C is a test setting that requires addressing all three targeted challenges simultaneously. Our objective was not to address each of the three challenges in isolation, but to establish an agent that can utilize the information embedded in commands to flexibly address the three challenges when required. Test C encompassed a wide range of situations. It included tasks both with and without attribute-based or contextual conditions, as well as various mid-execution commands such as task addition, removal, cancellation, reordering, and modifications to detailed task instructions. Achieving a high success rate (i.e., task success rate of 93.9\% and session success rate of 86.7\%) in real-world experiments with a physical robot under these diverse and dynamic conditions of Test C supports the robustness of the proposed system and demonstrates its ability to flexibly adapt its positioning strategy to improvisational instructions encountered in real construction environments.

The results from Test B particularly highlight that the robot can perform positioning while considering attribute and contextual conditions. Through the comparative analysis, we observed that prior open-vocabulary-based target identification approaches exhibit limitations in handling the complex contextual conditions evaluated in our tests. Within our agent framework, the LMM directly analyzes visual input and determines the robot’s movement, enabling the consideration of complex contextual and attribute-based conditions. This capability provides a foundation for construction robots to autonomously execute more complex improvisational tasks with attribute and contextual instructions in construction environments.

Across the three tests, a total of eight failed sessions occurred. Analyzing the causes of these failed sessions allows us to identify which aspects of the process were most challenging for the agent. In all eight cases, each failed session included one failed task, and there were no instances in which all tasks were successful, but the session itself failed. This suggests that the agent core robustly handled the processing and decomposition of natural language commands, as well as plan revisions in response to additional instructions introduced during execution. 

Table \ref{tab:6} analyzes the causes of task failures. Out of the eight total task failures, one occurred in the navigation module, while the remaining seven occurred in the positioning module. Most failures occurred because the robot did not move close enough to the task target, or because it mistakenly navigated to another object or a distractor, confusing it with the task target. The high number of failures in the positioning module is attributed to LMM’s relatively weaker performance in image-based reasoning and visual grounding compared to language-based reasoning \cite{ref_48}. In addition, the failure to reduce the distance to the target while jointly considering visual observations and front clearance reflects the limitations of LMMs in spatial reasoning from images \cite{ref_61}. To address this limitation, future work could explore improving the ability of LMMs to accurately understand the relative spatial relationships among the environment, the target, and the robot’s own position from the given field-of-view image.

The time required for the positioning may pose limitations when responding to time-sensitive improvisational tasks. Table \ref{tab:7} summarizes the robot’s total travel distance and time across the three tests, as well as the average travel distance and time per task. This time includes the entire process, including task breakdown, incorporating mid-execution commands, zone-level navigation, and vision-based positioning. In successful sessions, each task took approximately 2 minutes, whereas in failed sessions, each task took about 3 minutes. The longer duration in failed sessions was mainly due to the robot spending additional time searching for the target. The primary source of increased operation time is the repeated time required for LMM reasoning via API calls. Recent studies have attempted to reduce this delay by employing lightweight local language models that can be run on robot hardware \cite{ref_62}. However, lightweight language models exhibit lower reasoning capabilities than LMM, particularly for vision-based reasoning tasks involving contextual conditions \cite{ref_63}. Therefore, future research could investigate strategies that evaluate task complexity and selectively employ larger models when necessary.

Finally, testing in real construction sites has not been conducted. Compared to the environment used in our tests, real construction sites may contain many factors that act as distractors, which could make visually identifying task-required locations more challenging. In addition, during positioning, the robot must move more robustly by considering obstacles on site, as well as moving workers and equipment \cite{ref_32}. Therefore, future work should involve testing the agent in real construction site environments and ensuring that it can operate more robustly under such conditions.

\section{Conclusion}

This study presents a multi-LMM-module agent capable of understanding natural language task instructions, identifying task-required locations, and performing positioning to respond to improvisational tasks on construction sites. In particular, the agent is designed to address three key challenges inherent to improvisational tasks on construction: (1) work targets for which precise location coordinates are not specified in advance; (2) tasks that include attributes or contextual conditions that may be provided on site; and (3) the unexpected timing for task assignment or modification. Three tests were constructed to test the agent’s ability to address the three challenges. Across the three tests, task and session success rates reached 92.2\% and 82.2\%, respectively. In the final Test C, where all three challenges must be considered simultaneously, the agent achieved a task success rate of 93.9\% and a session success rate of 86.7\%. The proposed multi-LMM agent contributes to enabling mobile construction robots to perform improvisational tasks in construction environments where not all processes can be executed through predefined workflows.

\bibliography{Improvisational}

\end{document}